\pdfoutput=1

\documentclass[11pt]{article}

\usepackage[]{acl}

\usepackage{times}
\usepackage{latexsym}

\usepackage[T1]{fontenc}

\usepackage{graphicx}
\usepackage{booktabs}
\usepackage{multirow}
\usepackage{rotating}
\usepackage{amsmath}
\usepackage{amssymb}
\usepackage{bm}
\usepackage{framed}
\usepackage{caption}
\usepackage{xcolor}
\usepackage{algorithm}
\usepackage[noend]{algorithmic}
\usepackage{soul}
\usepackage{hyperref}
\usepackage{colortbl}
\usepackage{amssymb}
\usepackage{makecell}
\usepackage{bbding}
\usepackage{enumitem}

\definecolor{increase}{HTML}{FCE4D6}
\definecolor{decrease}{HTML}{D9E1F2}

\newcommand*\samethanks[1][\value{footnote}]{\footnotemark[#1]}
\newcommand{\tabincell}[2]{\begin{tabular}{@{}#1@{}}#2\end{tabular}}

\usepackage[utf8]{inputenc}

\usepackage{microtype}

\title{\textsc{CDConv}: A Benchmark for Contradiction Detection in \\ Chinese Conversations}

\author{
    Chujie Zheng$^1$\thanks{\ \ Equal contribution.} \quad Jinfeng Zhou$^{1,2}$\samethanks{} \quad Yinhe Zheng$^3$ \quad Libiao Peng$^3$ \quad Zhen Guo$^4$  \\
    \textbf{Wenquan Wu$^4$ \quad Zheng-Yu Niu$^4$ \quad Hua Wu$^4$ \quad Minlie Huang$^{1,3}$\thanks{\ \ Corresponding author.}} \\ 
    \small $^1$The CoAI Group, Institute for Artificial Intelligence, State Key Lab of Intelligent Technology and Systems, \\
    \small $^1$Beijing National Research Center for Information Science and Technology, DCST, Tsinghua University, Beijing 100084, China \\
    \small $^2$College of Intelligence and Computing, Tianjin University, Tianjin, China \\
    \small $^3$Lingxin AI, Beijing 100084, China \quad $^4$Baidu Inc., China \\ 
    \small \tt chujiezhengchn@gmail.com \quad jfzhou.mail@gmail.com \quad aihuang@tsinghua.edu.cn \\
    \small \tt \{guozhenguozhen, wuwenquan01, niuzhengyu, wu\_hua\}@baidu.com \\
}

\begin{document}

\maketitle
\begin{abstract}
    Dialogue contradiction is a critical issue in open-domain dialogue systems.
    The contextualization nature of conversations makes dialogue contradiction detection rather challenging.
    In this work, we propose a benchmark for \underline{\textbf{C}}ontradiction \underline{\textbf{D}}etection in Chinese \underline{\textbf{Conv}}ersations, namely \textbf{\textsc{CDConv}}.
    It contains 12K multi-turn conversations annotated with three typical contradiction categories: Intra-sentence Contradiction, Role Confusion, and History Contradiction.
    To efficiently construct the \textsc{CDConv} conversations, we devise a series of methods for automatic conversation generation, which simulate common user behaviors that trigger chatbots to make contradictions.
    We conduct careful manual quality screening of the constructed conversations and show that state-of-the-art Chinese chatbots can be easily goaded into making contradictions.
    Experiments on \textsc{CDConv} show that properly modeling contextual information is critical for dialogue contradiction detection, but there are still unresolved challenges that require future research.\footnote{Our data and codes are available at \url{https://www.github.com/thu-coai/CDConv} and \url{https://github.com/PaddlePaddle/Knover/tree/dygraph/projects/cdconv}}
\end{abstract}

\section{Introduction}
\label{sec:intro}

Large-scale pre-training for dialogue generation \cite{dialogpt, meena} has advanced the development of engaging and human-like dialogue systems.
Unfortunately, state-of-the-art open-domain chatbots, such as BlenderBot \cite{blenderbot}, EVA \cite{eva, eva2.0} and PLATO \cite{platoxl}, still often behave inconsistently with their role or identity and produce utterances that are self-contradictory or contradict the dialogue history \cite{identity, eva2.0, long-term-memory}.
Such inconsistency or contradiction phenomena violate Grice’s cooperative principle \cite{grice1975logic} and greatly impair the users' long-term trust \cite{huang2020challenges, blenderbot2.0-study}.

\begin{figure}[t]
    \centering
    \includegraphics[width=\linewidth]{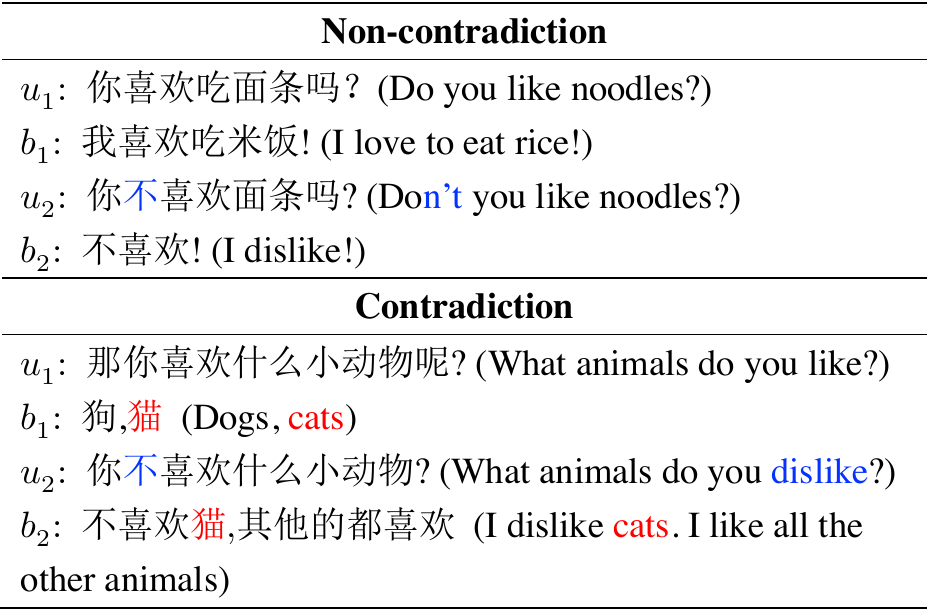}
    \caption{
    Dialogue contradiction detection requires the full contextual information (including $u_1$ and $u_2$) rather than only the bot's utterances (i.e., $b_1$ and $b_2$).
    }
    \label{fig:intro_example}
    \vspace{-1mm}
\end{figure}

\begin{table*}[t]
    \centering
    \scalebox{0.85}{
      \begin{tabular}{lcccc}
      \toprule
            & \textbf{Lang} & \textbf{Task Input} & \textbf{Task Type} & \textbf{Contradiction Categories} \\
      \midrule
      MNLI \citeyearpar{mnli} & En    & Sentence Pair & -     & - \\
      CMNLI \citeyearpar{clue}, OCNLI \citeyearpar{ocnli} & Zh    & Sentence Pair & -     & - \\
      DNLI \citeyearpar{dnli}, InferConvAI \citeyearpar{inferconvai} & En    & Sentence Pair & -     & - \\
      \midrule
      KvPI \citeyearpar{kvpi} & Zh    & Conversation \& Profile & Extrinsic & Profile \\
      \textsc{DialFact} \citeyearpar{dialfact} & En    & Conversation & Extrinsic & Fact \\
      CI-ToD \citeyearpar{ci-tod} & En    & Conversation \& KB & Int \& Ext & Query, History \& KB \\
      DECODE \citeyearpar{decode} & En    & Conversation & Intrinsic & History \\
      \midrule
      \textbf{\textsc{CDConv} (Ours)} & \textbf{Zh} & \textbf{Conversation} & \textbf{Intrinsic} & \textbf{Intra-sentence, Role, History} \\
      \bottomrule
      \end{tabular}%
      }
    \caption{
      Comparison of \textsc{CDConv} with related benchmarks / datasets for (dialogue) contradiction detection.
      The \textbf{Extrinsic} type targets the contradiction between a conversation and \textit{external information} (e.g., profiles or facts), while \textbf{Intrinsic} targets the contradiction \textit{inside} a conversation.
      See \S \ref{sec:related} for detailed discussion.
      }
    \label{tab:comparison}%
    \vspace{-1mm}
\end{table*}%

Dialogue contradiction detection has shown to be an effective means to improve the consistency of chatbots \cite{dnli, decode}, which, however, is always a challenging task.
Specifically, the contextualization nature of conversations indicates the necessity of considering and modeling contextual information.
For instance, in the ``Contradiction'' example in Figure \ref{fig:intro_example}, $b_2$ does not explicitly contradict $b_1$.
However, given $u_1$, the actual meaning of $b_1$ should be ``\underline{I like} dogs, cats'' and $b_1$ and $b_2$ are thus contradictory.
In contrast, in the ``Non-contradiction'' example, while $b_1$ and $b_2$ seem inconsistent (``love'' vs. ``dislike''), $b_2$ actually means ``I dislike \underline{noodles}'' considering the dialogue context.
Hence, $b_2$ is compatible with $b_1$ and does not make a contradiction.

Despite the above challenge, existing datasets for contradiction detection \cite{inferconvai, dnli} usually only consider the textual entailment relationship between two isolated sentences \cite{dagan2005pascal}, which is largely insufficient for dialogue contradiction detection due to the neglect of contextual information.
A recent work \cite{decode} crowd-sourced a dataset named DECODE that contains conversations where the last utterances contradict the dialogue histories.
However, DECODE lacks a wide coverage of typical contradiction categories, and most of its contradiction cases are written by human, which have gap with the real scenario where users trigger chatbots to make contradictions.

\begin{figure*}[t]
    \centering
    \includegraphics[width=\linewidth]{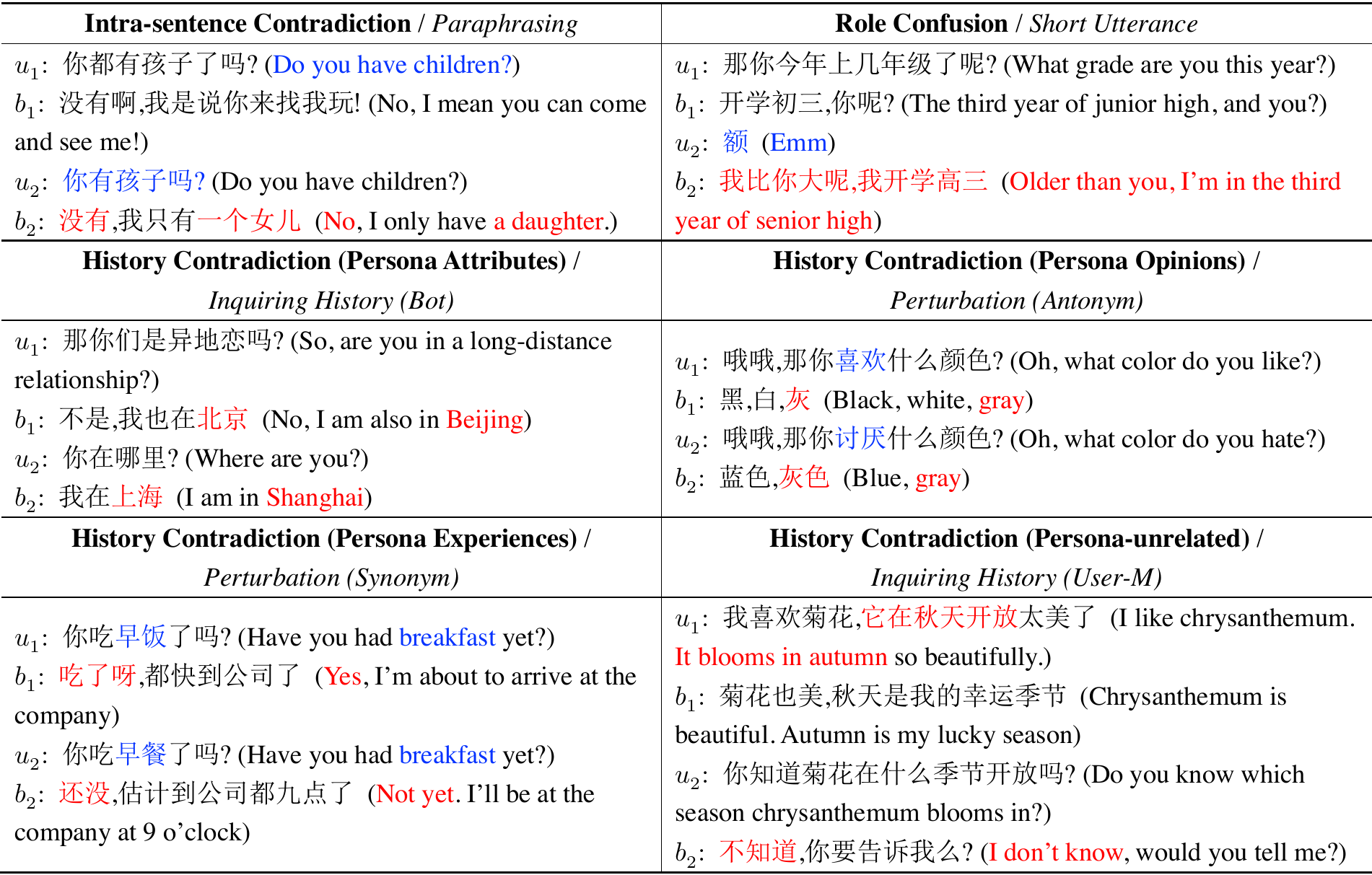}
    \caption{
    Data examples of \textbf{contradiction categories (\S \ref{sec:categories})} / \textit{trigger methods (\S \ref{subsec:methods})}.
    \textcolor{red}{Red texts} denote the parts that make contradiction.
    \textcolor{blue}{Blue texts} illustrate the \textit{trigger methods} (i.e., how $u_2$ are constructed).
    \textit{Perturbation (Negative)} and \textit{Inquiring History} are separately illustrated in Figure \ref{fig:intro_example} and Figure \ref{fig:inquiring} respectively.
    }
    \label{fig:data_example}
    \vspace{-2mm}
\end{figure*}

In this work, we propose a benchmark for \underline{\textbf{C}}ontradiction \underline{\textbf{D}}etection in Chinese \underline{\textbf{Conv}}ersations, namely \textbf{\textsc{CDConv}}.
It contains 12K multi-turn conversations with human-annotated contradiction labels (\S \ref{sec:categories}).
Different from previous work (e.g., \citealt{decode}) that only considered the contradiction to \textit{dialogue history} (i.e., History Contradiction), \textsc{CDConv} covers another two typical categories: Intra-sentence Contradiction and Role Confusion, which refer to that a reply contradicts \textit{itself} and that a reply confuses \textit{the speaker's role}, respectively.

Since the cases of non-contradiction and contradiction in natural human-bot conversations are extremely unbalanced (\S \ref{sec:categories}, \citealt{decode}), we automatically construct the \textsc{CDConv} conversations combined with elaborate manual inspection (\S \ref{subsec:procedure}).
Specifically, we first devise a series of automatic methods to generate conversations (\S \ref{subsec:methods}), which simulate the common user behaviors that trigger chatbots to make contradictions.
We then conduct careful human screening and annotation for the constructed conversations to ensure the data quality (\S \ref{subsec:quality}).
We validate the effectiveness of the trigger methods and show that state-of-the-art Chinese open-domain chatbots (EVA and PLATO) can be easily goaded into making contradictions (\S \ref{subsec:statistics}).

Finally, we evaluate popular Chinese pre-trained models on \textsc{CDConv} (\S \ref{sec:experiment}).
Results show that properly modeling contextual information is critical for dialogue contradiction detection.
However, there is still much room for future research in dialogue modeling, integrating commonsense and world knowledge, and reasoning.

Our contributions are summarized as follows:

\begin{itemize}[leftmargin=*]

    \item We propose \textsc{CDConv}, a benchmark for contradiction detection in Chinese conversations.
    It contains 12K conversations annotated with three typical contradiction categories: Intra-sentence Contradiction, Role Confusion, and History Contradiction.
    
    \item We present a series of methods by simulating common user behaviors to automatically trigger chatbots to make contradictions.
    We demonstrate the effectiveness of these trigger methods through detailed human annotation.
    
    \item We evaluate popular Chinese pre-trained models on \textsc{CDConv}.
    Results show the importance of properly modeling contextual information in dialogue contradiction detection, while this task is still far from solved and requires further study.

\end{itemize}

\section{Related Work}
\label{sec:related}

Table \ref{tab:comparison} summarizes the comparison of \textsc{CDConv} with related benchmarks / datasets for (dialogue) contradiction detection.

\vspace{-1mm}
\paragraph{Contradiction Detection for Sentence Pair}
The early contradiction detection usually adopted the natural language inference (NLI) framework \cite{dagan2005pascal}, such as the English MNLI \cite{mnli} dataset and the Chinese CMNLI \cite{clue} and OCNLI \cite{ocnli} datasets.
The task input consists of two isolated sentences, which are labeled as one of the textual entailment relationships: ``entailment'', ``neutral'' and ``contradiction''.
To extend the NLI framework to the dialogue domain, \citet{dnli} constructed the DNLI dataset where the dialogue utterances and the persona descriptions from PersonaChat \cite{pc} are used to form sentence pairs.
\citet{inferconvai} similarly synthesized the InferConvAI dataset through automatic manipulation with dialogue utterances.
However, the NLI framework does not consider the contextualization nature of conversations, making it deficient for dialogue contradiction detection.

\vspace{-1mm}
\paragraph{Contradiction Detection for Conversation}
The contradictions in dialogue systems can be split into two major types: Extrinsic and Intrinsic \cite{path-hunter, ji2022survey}.
The \textbf{Extrinsic} type refers to the contradiction between a conversation and \textit{external information}.
For instance, the KvPI dataset \cite{kvpi} focuses on the contradiction to structured attribute profiles.
The \textsc{DialFact} benchmark \cite{dialfact} aims at detecting contradictory statements to world facts and improving factual correctness.
The CI-ToD dataset \cite{ci-tod} involves the inconsistency with knowledge bases in task-oriented dialogue.
One potential limitation of Extrinsic dialogue contradiction detection is that it may rely on static and manually curated external information (e.g., profiles), which could be insufficient in open-domain dialogue.

Our work focuses on the \textbf{Intrinsic} type, which refers to the contradiction \textit{inside} a conversation and is more widespread and fundamental in open-domain dialogue.
The DECODE dataset \cite{decode} is a relevant work to ours, whose contradiction cases are mostly collected by manually writing subsequent utterances to contradict the given dialogue histories.
Besides the language difference, \textsc{CDConv} is distinguished from DECODE in two aspects: 
(1) Apart from History Contradiction, \textsc{CDConv} additionally covers two contradiction categories: Intra-sentence Contradiction and Role Confusion, which are also typical and common in human-bot conversations (\S \ref{sec:categories}).
(2) Instead of being human-written, the contradiction cases in \textsc{CDConv} are constructed by simulating the user behaviors that trigger chatbots to make contradictions (\S \ref{subsec:methods}), which are closer to the real scenario of human-bot conversation.

\section{Categories of Dialogue Contradiction}
\label{sec:categories}

A conversation with $n$ turns is formally denoted as $u_1, b_1, \dots, u_n, b_n$, where $u_k$ and $b_k$ denote the $k$th-turn utterances from the user and the chatbot respectively.
We focus on whether $b_n$ makes a contradiction in the dialogue context.

In the preliminary study, we manually inspected 200 multi-turn human-bot conversations with two Chinese open-domain chatbots: EVA \cite{eva, eva2.0} and PLATO \cite{plato2, platoxl}.
On average, each conversation contains about 30 turns but only roughly 1 contradiction case.
Based on the inspected contradiction cases, we identify three typical categories of dialogue contradiction according to \textit{the object that $b_n$ contradicts}, as intuitively illustrated by Figure \ref{fig:diagram}:

\begin{figure}[t]
    \centering
    \includegraphics[width=0.85\linewidth]{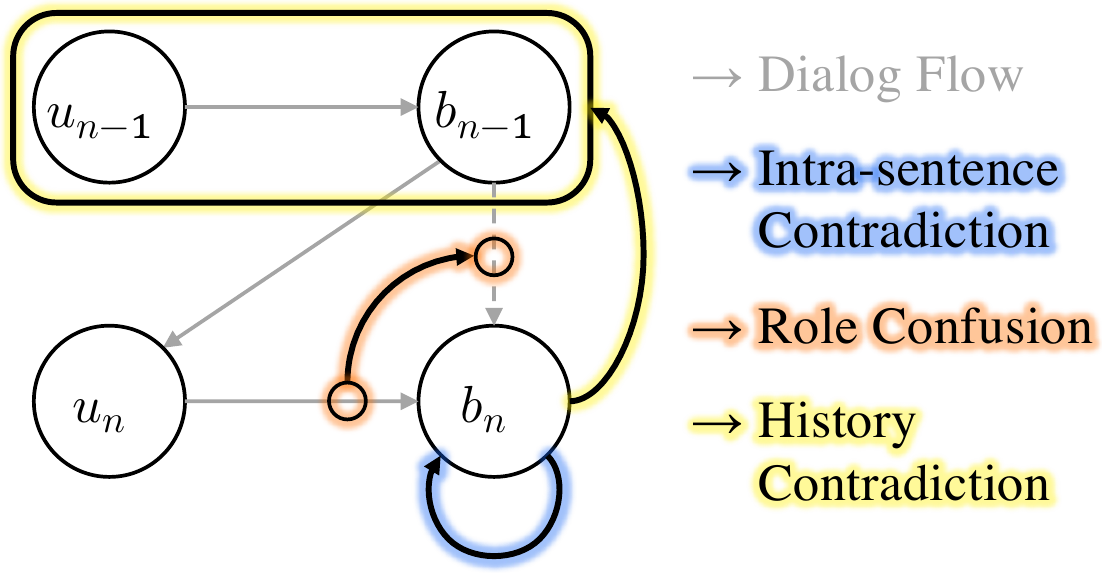}
    \caption{
    Diagram of contradiction categories.
    Combine the definitions below for a clearer understanding.
    }
    \label{fig:diagram}
    \vspace{-0mm}
\end{figure}

\begin{itemize}[leftmargin=*]

    \item \textbf{Intra-sentence Contradiction}:
    $b_n$ is contradictory to \textit{itself}.
    In other words, there exist two disjoint subsentences $b_n^{(1)}, b_n^{(2)} \subset b_n$ (usually separated by commas, periods or conjunctions) so that they are not compatible with each other.
    
    \item \textbf{Role Confusion}:
    $b_n$ confuses \textit{the speaker's role}.
    That is, $b_n$ is more likely to be a user's reply to $b_{n-1}$ rather than a bot's to $u_n$.
    
    \item \textbf{History Contradiction}\footnote{
    We note that the premise of $b_n$ making History Contradiction is that $b_n$ is a bot's reply to $u_n$.
    However, if $b_n$ makes Role Confusion (i.e., $b_n$ is more likely to be a user's reply to $b_{n-1}$ than a bot's reply to $u_n$), the premise of History Contradiction will not hold and such a case will be judged as Role Confusion rather than History Contradiction.
    }:
    $b_n$ is contradictory to \textit{the dialogue history}.
    The contradictions caused by mistaking or forgetting the dialogue history \cite{long-term-memory, dulemon} usually fall into History Contradiction, as the last example in Figure \ref{fig:data_example}.

\end{itemize}

Figure \ref{fig:data_example} provides the examples of the above three contradiction categories.
They occupied 16\%, 18\%, and 54\% in our inspected contradiction cases, respectively.
The remaining cases ($<$ 12\%) mostly contradict time-sensitive information (e.g., the chat time) or facts (e.g., when the iPhone was released), which, as aforementioned (\S \ref{sec:related}), are beyond the scope of this work.
We note that Intra-sentence Contradiction and Role Confusion were less studied previously while actually typical and common in human-bot conversations.
\textsc{CDConv} can serve as a good start point for investigating them.

\section{Data Collection}
\label{sec:collection}

\subsection{Collection Procedure}
\label{subsec:procedure}

We automatically constructed the \textsc{CDConv} conversations along with elaborate manual inspection.
We narrow down the conversations in \textsc{CDConv} to 2-turn ones ($n=2$).
The overview procedure is shown in Figure \ref{fig:procedure}:

\begin{figure}[t]
    \centering
    \includegraphics[width=\linewidth]{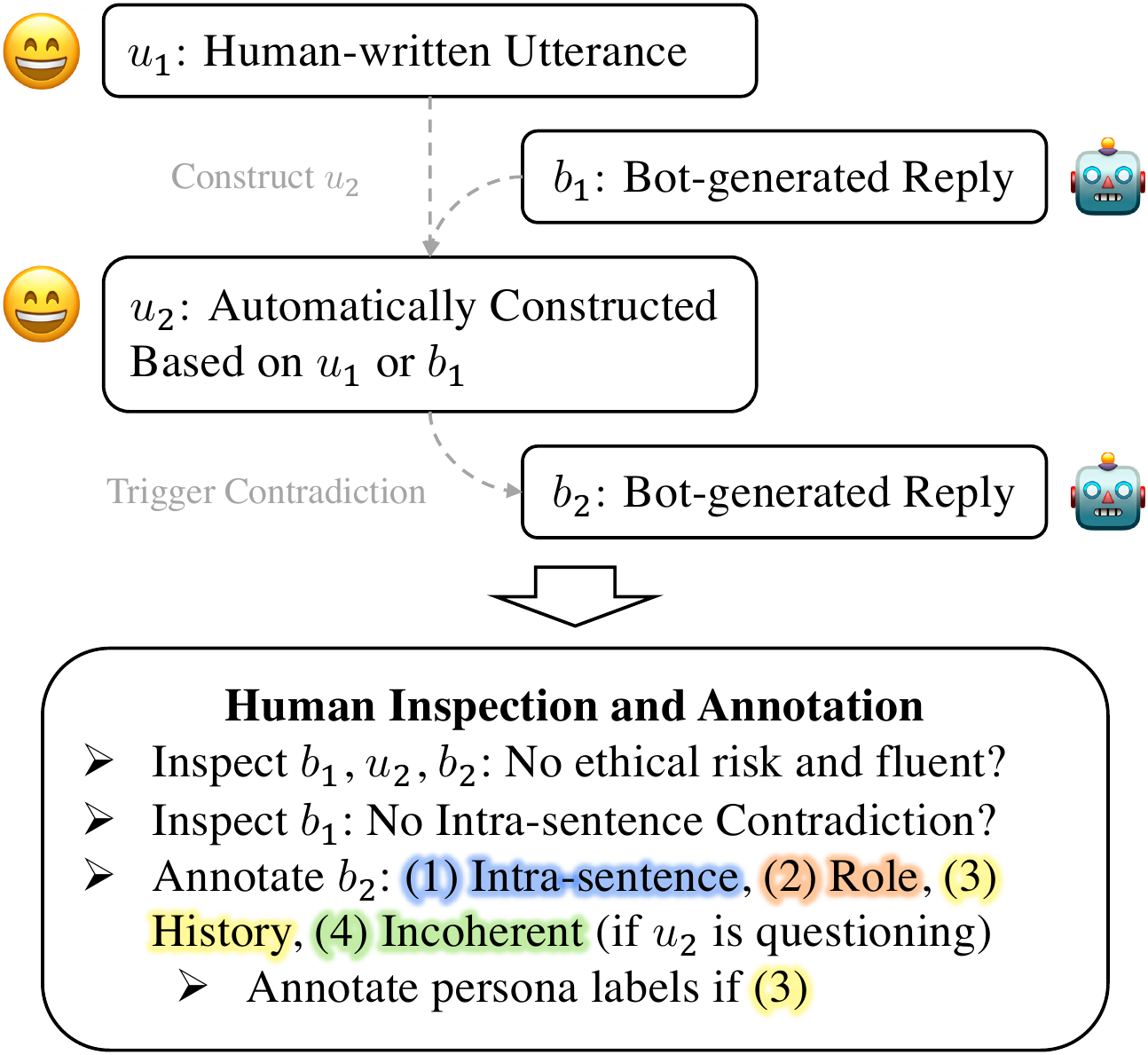}
    \caption{
    The collection procedure of \textsc{CDConv}.
    See Table \ref{tab:annotation} for detailed annotation statistics.
    }
    \label{fig:procedure}
    \vspace{-0mm}
\end{figure}

\begin{table*}[t]
    \centering
    \scalebox{0.85}{
      \begin{tabular}{lc||ccccc||ccccc}
      \toprule
      \multirow{3}[4]{*}{\textbf{Methods}} & \multicolumn{1}{c}{\multirow{3}[4]{*}{\tabincell{c}{\textbf{$\bm{u_2}$ Not} \\ \textbf{Fluent}}}} & \multicolumn{5}{c}{\textbf{EVA}} & \multicolumn{5}{c}{\textbf{PLATO}} \\
      \cmidrule(lr){3-7} \cmidrule(lr){8-12}      
      & \multicolumn{1}{c}{} & \multirow{2}[2]{*}{\tabincell{c}{\textbf{$\bm{b_1}$} \\ \textbf{Intra}}} & \multicolumn{4}{c}{\textbf{$\bm{b_2}$}} & \multirow{2}[2]{*}{\tabincell{c}{\textbf{$\bm{b_1}$} \\ \textbf{Intra}}} & \multicolumn{4}{c}{\textbf{$\bm{b_2}$}} \\
      \cmidrule(lr){4-7} \cmidrule(lr){9-12} 
      & \multicolumn{1}{c}{} &    &  \textbf{Intra} & \textbf{Role} & \textbf{History} &  \multicolumn{1}{c}{\textbf{Incoh}} &  &  \textbf{Intra} & \textbf{Role} & \textbf{History} &  \textbf{Incoh} \\
      \midrule
      Short & -  & \cellcolor[rgb]{ .831,  .898,  .961}0.04  & 0.00  & \cellcolor[rgb]{ .98,  .843,  .757}0.14  & \cellcolor[rgb]{ 1,  .996,  .976}0.04  & \cellcolor[rgb]{ 1,  1,  .996}0.00  & \cellcolor[rgb]{ .953,  .973,  .988}0.01  & \cellcolor[rgb]{ .973,  .98,  .992}0.01  & \cellcolor[rgb]{ .957,  .69,  .518}0.27  & \cellcolor[rgb]{ 1,  .996,  .984}0.03  & \cellcolor[rgb]{ 1,  1,  .996}0.00  \\
      Inquiring (Bot) & \cellcolor[rgb]{ .843,  .867,  .902}0.19  & \cellcolor[rgb]{ .608,  .761,  .902}0.08  & \cellcolor[rgb]{ .604,  .698,  .875}0.09  & \cellcolor[rgb]{ 1,  .984,  .976}0.02  & \cellcolor[rgb]{ 1,  .953,  .804}0.31  & \cellcolor[rgb]{ .949,  .973,  .929}0.03  & \cellcolor[rgb]{ .843,  .906,  .961}0.03  & \cellcolor[rgb]{ .886,  .914,  .965}0.03  & \cellcolor[rgb]{ .988,  .89,  .827}0.10  & \cellcolor[rgb]{ 1,  .973,  .89}0.17  & \cellcolor[rgb]{ .875,  .933,  .835}0.08  \\
      Inquiring (User) & \cellcolor[rgb]{ .871,  .89,  .918}0.16  & \cellcolor[rgb]{ .8,  .878,  .953}0.04  & \cellcolor[rgb]{ .867,  .898,  .961}0.03  & \cellcolor[rgb]{ .992,  .937,  .902}0.06  & \cellcolor[rgb]{ 1,  .953,  .8}0.31  & \cellcolor[rgb]{ .749,  .863,  .667}0.16  & \cellcolor[rgb]{ .973,  .984,  .996}0.01  & \cellcolor[rgb]{ .945,  .957,  .984}0.01  & \cellcolor[rgb]{ .984,  .867,  .788}0.12  & \cellcolor[rgb]{ 1,  .969,  .859}0.22  & \cellcolor[rgb]{ .663,  .816,  .557}0.22  \\
      Inquiring (User-M) & \cellcolor[rgb]{ .894,  .914,  .933}0.13  & \cellcolor[rgb]{ .886,  .929,  .973}0.02  & \cellcolor[rgb]{ .753,  .812,  .922}0.06  & \cellcolor[rgb]{ 1,  .996,  .996}0.00  & \cellcolor[rgb]{ 1,  .902,  .6}0.62  & \cellcolor[rgb]{ .98,  .992,  .973}0.01  & \cellcolor[rgb]{ .937,  .965,  .984}0.01  & \cellcolor[rgb]{ .894,  .918,  .969}0.03  & \cellcolor[rgb]{ .996,  .961,  .941}0.03  & \cellcolor[rgb]{ 1,  .933,  .725}0.43  & \cellcolor[rgb]{ .867,  .929,  .824}0.09  \\
      Paraphrasing & \cellcolor[rgb]{ .953,  .961,  .973}0.06  & \cellcolor[rgb]{ .729,  .835,  .933}0.06  & \cellcolor[rgb]{ .694,  .769,  .902}0.07  & \cellcolor[rgb]{ 1,  .992,  .988}0.01  & \cellcolor[rgb]{ 1,  .965,  .847}0.24  & \cellcolor[rgb]{ .996,  1,  .992}0.00  & \cellcolor[rgb]{ .918,  .953,  .98}0.02  & \cellcolor[rgb]{ .922,  .941,  .976}0.02  & \cellcolor[rgb]{ .992,  .922,  .875}0.07  & \cellcolor[rgb]{ 1,  .969,  .867}0.21  & \cellcolor[rgb]{ .922,  .957,  .898}0.05  \\
      Perturb (Synonym) & \cellcolor[rgb]{ .82,  .847,  .886}0.22  & \cellcolor[rgb]{ .745,  .847,  .937}0.05  & \cellcolor[rgb]{ .671,  .749,  .898}0.08  & \cellcolor[rgb]{ 1,  1,  .996}0.00  & \cellcolor[rgb]{ 1,  .961,  .843}0.25  & \cellcolor[rgb]{ .965,  .98,  .949}0.02  & \cellcolor[rgb]{ .906,  .941,  .976}0.02  & \cellcolor[rgb]{ .937,  .953,  .98}0.02  & \cellcolor[rgb]{ .996,  .945,  .914}0.05  & \cellcolor[rgb]{ 1,  .973,  .882}0.18  & \cellcolor[rgb]{ .804,  .894,  .741}0.13  \\
      Perturb (Antonym) & \cellcolor[rgb]{ .675,  .725,  .792}0.39  & \cellcolor[rgb]{ .733,  .839,  .933}0.06  & \cellcolor[rgb]{ .671,  .749,  .898}0.08  & \cellcolor[rgb]{ 1,  .992,  .988}0.01  & \cellcolor[rgb]{ 1,  .949,  .792}0.32  & \cellcolor[rgb]{ .894,  .945,  .863}0.07  & \cellcolor[rgb]{ .937,  .961,  .984}0.01  & \cellcolor[rgb]{ .855,  .89,  .953}0.03  & \cellcolor[rgb]{ .996,  .969,  .949}0.03  & \cellcolor[rgb]{ 1,  .976,  .902}0.16  & \cellcolor[rgb]{ .851,  .918,  .804}0.10  \\
      Perturb (Negative) & \cellcolor[rgb]{ .749,  .788,  .839}0.31  & \cellcolor[rgb]{ .776,  .863,  .945}0.05  & \cellcolor[rgb]{ .557,  .663,  .859}0.10  & \cellcolor[rgb]{ 1,  .996,  .992}0.01  & \cellcolor[rgb]{ 1,  .957,  .82}0.28  & \cellcolor[rgb]{ .957,  .976,  .945}0.03  & \cellcolor[rgb]{ .922,  .953,  .98}0.02  & \cellcolor[rgb]{ .847,  .886,  .953}0.04  & \cellcolor[rgb]{ .996,  .957,  .929}0.04  & \cellcolor[rgb]{ 1,  .976,  .902}0.15  & \cellcolor[rgb]{ .878,  .933,  .843}0.08  \\
      \midrule
      Macro-Average & 0.21  & 0.05  & 0.06  & 0.03  & 0.30  & 0.04  & 0.02  & 0.02  & 0.09  & 0.19  & 0.09  \\
      \bottomrule
      \end{tabular}%
      }
    \caption{
    Annotation statistics for each trigger method.
    Each value means the proportion of the corresponding annotation label.
    The proportions about $b_2$ are calculated after the unqualified conversations were filtered out (in the 3rd step in \S \ref{subsec:procedure}).
    The proportions of ethical risk and non-fluent $b_1, b_2$ are omitted since they are all close to 0.
    }
    \label{tab:annotation}%
\end{table*}%

\definecolor{non-fluent}{rgb}{ .675,  .725,  .792}
\definecolor{intra_b1}{rgb}{ .608,  .761,  .902}
\definecolor{intra_b2}{rgb}{ .557,  .663,  .859}
\definecolor{role}{rgb}{ .957,  .69,  .518}
\definecolor{history}{rgb}{ 1,  .902,  .6}
\definecolor{incoherent}{rgb}{ .663,  .816,  .557}
  
\begin{enumerate}[leftmargin=*]

    \item We took a human-written utterance as $u_1$ and obtained the chatbot's reply $b_1$.
    
    \item Using one of the trigger methods in \S \ref{subsec:methods}, we automatically constructed $u_2$ based on $u_1$ or $b_1$ and generated the chatbot's next reply $b_2$.
    
    \item Human annotators were asked to inspect (1) if $b_1, u_2, b_2$ do not contain any ethical risk (e.g., offensive language, hate speech, unethical suggestions, etc.) and are fluent and understandable, and (2) if $b_1$ does not make Intra-sentence Contradiction (to ensure a valid dialogue history).
    The unqualified conversations were removed.
    
    \item Considering the full contextual information, human annotators marked whether $b_2$ makes a contradiction based on the categories in \S \ref{sec:categories}.
    Specifically, we adopted single-label annotation. 
    That is, according to the order in \S \ref{sec:categories}, once a contradiction of some category is recognized, the subsequent categories will not be judged.
    Note that the cases, where $b_2$ does not answer the questioning $u_2$ and responds incoherently (e.g., unnaturally transition the topic), were additionally marked and filtered out.

\end{enumerate}

\vspace{-1mm}
\paragraph{Collecting $\bm{u_1}$}
We collected the human-written utterances from DuPersona, a crowd-sourced Chinese open-domain dialogue corpus\footnote{\url{https://www.luge.ai/\#/luge/dataDetail?id=38}}.
This is due to our observation that these crowd-sourced utterances are of higher quality compared to social media posts (e.g., Weibo and Douban) and contain rich persona information, which is in line with the style and content of general chitchat.
We used those utterances that contain second-person nouns and ``?'' as $u_1$, since noticed that such questioning utterances would elicit chatbots to talk specific information about themselves and could avoid uninformative or meaningless replies.

\vspace{-1mm}
\paragraph{Persona Labels}
To help understand which type of information was involved in History Contradiction, these $b_2$ were additionally annotated with one of the four persona labels: attributes, opinions, experiences and persona-unrelated.
Their examples are shown in Figure \ref{fig:data_example} and their definitions are provided in \S \ref{sec:persona_definition}.
Note that we annotated the persona information since its related discussion in Chinese chitchat usually occupies a large proportion according to our observations on social media corpora.

\vspace{-1mm}
\paragraph{Chatbots} 
We used two state-of-the-art Chinese open-domain chatbots, EVA \cite{eva, eva2.0} and PLATO \cite{plato2, platoxl}.
EVA is an Encoder-Decoder model with 24 encoder layers and 24 decoder layers and has 2.8B parameters in total.
PLATO adopts a Unified Transformer architecture \cite{plato} and has 32 layers and 1.6B parameters.
They are both pre-trained on massive Chinese social media corpora.

\subsection{Trigger Methods}
\label{subsec:methods}

\begin{figure}[t]
    \centering
    \includegraphics[width=\linewidth]{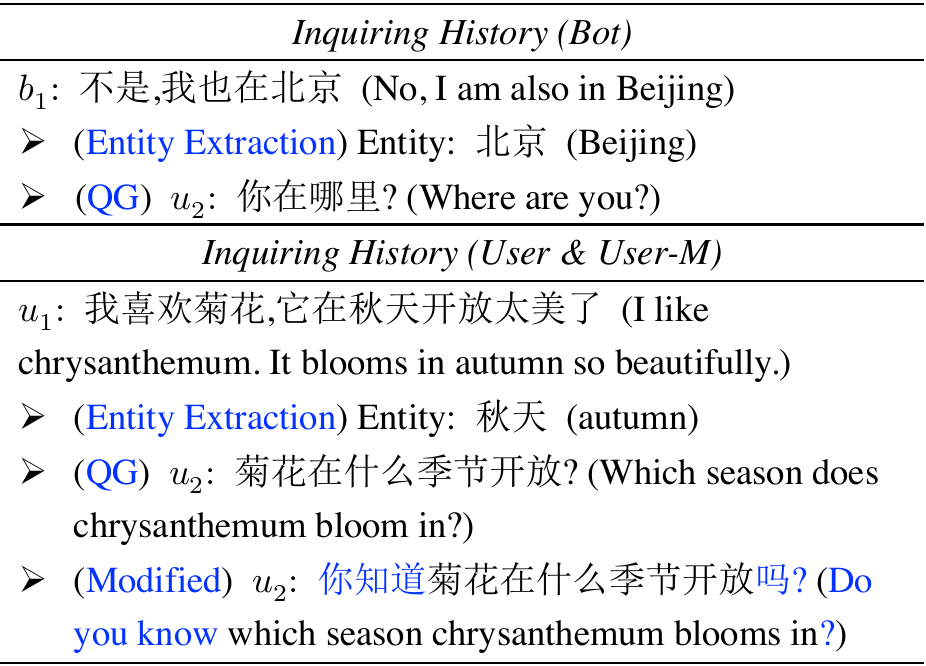}
    \caption{
    Illustration of \textit{Inquiring History}.
    }
    \label{fig:inquiring}
    \vspace{-0mm}
\end{figure}

Our inspection on contradiction cases (\S \ref{sec:categories}) also revealed that chatbots are more prone to making contradictions under several specific user behaviors: (1) the user input is short and uninformative, (2) the user inquires about the dialogue history (similarly noticed by \citealt{inquiry-history}), and (3) the user asks for similar information in the context.
By simulating these user behaviors, we devise a series of methods to automatically construct $u_2$.
These methods are illustrated by the examples in Figure \ref{fig:intro_example}, \ref{fig:data_example} and \ref{fig:inquiring}.
Note that the automatic construction of $u_2$ suggests the necessity of inspecting if it is fluent and understandable, which is thus an important step to ensure data quality (\S \ref{subsec:procedure}).

\vspace{-1mm}
\paragraph{Short Utterance}
$u_2$ is a short and uninformative utterance.
It simulates a user's casual or perfunctory reply to the chatbot.

With manual screening, we collected 145 short utterances ($\le 3$ characters) from DuPersona as $u_2$.

\vspace{-1mm}
\paragraph{Inquiring History (Bot / User)}
$u_2$ is an inquiry about the dialogue history.
It simulates a user's inquiry about the contents of previous conversations.

We first extracted named entities in $b_1$ (about the bot) or $u_1$ (about the user) using HanLP\footnote{\url{https://github.com/hankcs/HanLP}}
\cite{hanlp}.
Then we leveraged an open-sourced question generation model\footnote{\url{https://github.com/artitw/text2text}} to generate questions about the extracted entities, which were used as $u_2$.

Note that when inquiring about the user, we used the utterances that contain first-person nouns from DuPersona as $u_1$.
Since we noticed that such obtained $u_2$ was sometimes not natural enough, we modified most of $u_2$ using the pattern ``Do you know...?'', which we denote as \textbf{Inquiring History (User-M)}, as illustrated in Figure \ref{fig:inquiring}.

\vspace{-1mm}
\paragraph{Paraphrasing}
$u_2$ expresses the same meaning to $u_1$ in a different way.
It simulates a user's clarification question to the previous questions.

We paraphrased $u_1$ through back-translation as $u_2$.
The Chinese $u_1$ was first translated to English and then back-translated to Chinese.
We used the Baidu translation API and removed those $u_2$ that were identical to $u_1$.

\vspace{-1mm}
\paragraph{Perturbation}
As an extension of Paraphrasing, we found that $u_2$ obtained by perturbing $u_1$, where $u_2$ and $u_1$ have similar or opposite meanings, could also trigger contradictions.
Different from the methods before, Perturbation is more likely to be users' ``hacking'' behaviors instead of general chitchat, which may be out of the intents of curiosity, probing, or malicious attacks, etc.

We perturbed $u_1$ in three ways.
(1) \textbf{Synonym}.
We randomly replaced the nouns in $u_1$ with their synonyms using an open-sourced synonym dictionary\footnote{\url{https://github.com/guotong1988/chinese\_dictionary}}.
(2) \textbf{Antonym}.
We randomly replaced the verbs or adjectives in $u_1$ with their antonyms using the antonym dictionary.
For Synonym and Antonym, there are 2.3/3.7 words per $u_1$ on average that can be replaced with their synonyms/antonyms.
In practice, we randomly chose one replaceable word in $u_1$ at a time.
(3) \textbf{Negative}.
We randomly replaced the words in $u_1$ with their negatives using the negative dictionary or inserted negatives before the verbs in $u_1$.
Since we noticed that negatives would greatly impair the fluency of $u_2$, we additionally applied back-translation to $u_2$ to improve its fluency.

\subsection{Quality Control}
\label{subsec:quality}

All the human annotators were hired from a reputable data annotation company.
They were instructed with the annotation procedure and the definitions and examples of contradiction categories.
However, due to the characteristics of the Chinese language and the difference in individual habits of language usage and communication, the annotation criteria of the annotators may somewhat vary and need to be calibrated with our assistance.
We applied the following mechanisms for quality control:

\vspace{-1mm}
\paragraph{Annotator Training}
All the annotators were required to take a training tutorial, which consists of 50 conversations for pilot annotation.
We provided feedback to help them calibrate the annotation criteria.

\vspace{-1mm}
\paragraph{Multi-person Annotation}
In the formal annotation, each conversation was annotated by two different annotators. 
If their results were inconsistent, a third annotator would be asked to re-annotate and discuss the case with the first two annotators to reach a consensus.

\vspace{-1mm}
\paragraph{Spot Check}
To more effectively calibrate the annotation criteria, we conducted annotation batch by batch and randomly sampled 100 conversations each batch for spot check.
We provided feedback to the annotators and instructed them to amend their annotations.
After each revision we would conduct spot check again until the pass rate reached 95\%.
Finally, we conducted five batches of annotation with incremental batch sizes (17K annotated conversations in total).
Except for the first two batches, all subsequent batches directly passed the first spot checks.

\begin{table}[t]
    \centering
    \scalebox{0.85}{
      \begin{tabular}{lrrr}
      \toprule
         & \textbf{EVA} & \textbf{PLATO} & \textbf{Total} \\
      \midrule
      \# Conversations & 5,458  & 6,202  & 11,660  \\
      \# Positive & 3,233  & 4,076  & 7,309  \\
      \# Negative & 2,225  & 2,126  & 4,351  \\
      \midrule
      \multicolumn{4}{l}{\textit{Trigger Methods (Positive / Negative Samples)}} \\
      \small \# Short & \small 429 / 91 & \small 692 / 304 & \small 1,121 / 395 \\
      \small \# Inquiring (Bot) & \small 764 / 577 & \small 845 / 406 & \small 1,609 / 983 \\
      \small \# Inquiring (User) & \small 127 / 116 & \small 131 / 106 & \small 258 / 222 \\
      \small \# Inquiring (User-M) & \small 251 / 552 & \small 477 / 541 & \small 728 / 1,093 \\
      \small \# Paraphrasing & \small 962 / 448 & \small 846 / 389 & \small 1,808 / 837 \\
      \small \# Perturb (Synonym) & \small 288 / 145 & \small 376 / 147 & \small 664 / 292 \\
      \small \# Perturb (Antonym) & \small 185 / 143 & \small 319 / 103 & \small 504 / 246 \\
      \small \# Perturb (Negative) & \small 227 / 153 & \small 390 / 130 & \small 617 / 283 \\
      \midrule
      \multicolumn{4}{l}{\textit{Contradiction Categories (of Negative Samples)}} \\
      Intra-sentence & 17.3\% & 6.8\% & 12.2\% \\
      Role & 5.8\% & 29.9\% & 17.6\% \\
      History & 76.9\% & 63.3\% & 70.2\% \\
      \midrule
      \multicolumn{4}{l}{\textit{Persona Labels (of History Contradiction)}} \\
      Attributes & 48.8\% & 46.2\% & 47.7\% \\
      Opinions & 22.2\% & 20.7\% & 21.5\% \\
      Experiences & 26.3\% & 31.5\% & 28.6\% \\
      Unrelated & 2.7\% & 1.6\% & 2.2\% \\
      \bottomrule
      \end{tabular}%
      }
    \caption{Statistics of \textsc{CDConv}.}
    \label{tab:dataset}%
    \vspace{-0mm}
\end{table}%

\begin{table*}[t]
    \centering
    \scalebox{0.85}{
      \begin{tabular}{cccc||cc||cccc}
      \toprule
      \multirow{2}[2]{*}{\textbf{Models}} & \multirow{2}[2]{*}{\textbf{Methods}} & \multicolumn{2}{c}{\textbf{2-class}} & \multicolumn{2}{c}{\textbf{4-class}} & \multicolumn{4}{c}{\textbf{4-class (Fine-grained F1)}} \\
      \cmidrule(lr){3-4}\cmidrule(lr){5-6}\cmidrule(lr){7-10}
      &    & \textbf{Acc} & \multicolumn{1}{c}{\textbf{F1}} & \textbf{Acc} & \multicolumn{1}{c}{\textbf{F1}} & \textbf{Non} & \textbf{Intra} & \textbf{Role} & \textbf{History} \\
      \midrule
      \multirow{5}[0]{*}{BERT} & Sentence Pair & 75.3 & 73.8 & 72.3 & 54.5 & 81.0 & 24.0 & 48.5 & 64.4 \\
         & \multirow{2}[0]{*}{Flatten} & 77.6 & 75.8 & 73.6 & 54.6 & 81.8 & 28.5 & 38.8 & 69.1 \\
         &    & \colorbox{increase}{\footnotesize +2.3} & \colorbox{increase}{\footnotesize +2.0} & \colorbox{increase}{\footnotesize +1.3} & \colorbox{increase}{\footnotesize +0.1} & \colorbox{increase}{\footnotesize +0.8} & \colorbox{increase}{\footnotesize +4.6} & \colorbox{decrease}{\footnotesize -9.7} & \colorbox{increase}{\footnotesize +4.7} \\
         & \multirow{2}[0]{*}{Hierarchical} & 77.9 & 75.9 & 75.2 & 56.6 & 83.1 & 30.0 & 44.2 & 68.9 \\
         &    & \colorbox{increase}{\footnotesize +2.6} & \colorbox{increase}{\footnotesize +2.1} & \colorbox{increase}{\footnotesize +3.0} & \colorbox{increase}{\footnotesize +2.1} & \colorbox{increase}{\footnotesize +2.1} & \colorbox{increase}{\footnotesize +6.0} & \colorbox{decrease}{\footnotesize -4.3} & \colorbox{increase}{\footnotesize +4.5} \\
      \midrule
      \multirow{5}[0]{*}{RoBERTa} & Sentence Pair & 75.7 & 73.7 & 72.2 & 55.1 & 81.2 & 29.1 & 46.5 & 63.4 \\
         & \multirow{2}[0]{*}{Flatten} & 78.6 & 77.0 & 75.7 & 56.8 & 84.1 & 28.8 & 43.3 & 70.9 \\
         &    & \colorbox{increase}{\footnotesize +2.9} & \colorbox{increase}{\footnotesize +3.2} & \colorbox{increase}{\footnotesize +3.4} & \colorbox{increase}{\footnotesize +1.7} & \colorbox{increase}{\footnotesize +2.8} & \colorbox{decrease}{\footnotesize -0.3} & \colorbox{decrease}{\footnotesize -3.2} & \colorbox{increase}{\footnotesize +7.5} \\
         & \multirow{2}[0]{*}{Hierarchical} & \textbf{80.4} & \textbf{78.1} & \textbf{77.8} & \textbf{59.3} & \textbf{85.1} & \textbf{33.0} & 48.1 & \textbf{71.0} \\
         &    & \colorbox{increase}{\footnotesize +4.7} & \colorbox{increase}{\footnotesize +4.4} & \colorbox{increase}{\footnotesize +5.5} & \colorbox{increase}{\footnotesize +4.3} & \colorbox{increase}{\footnotesize +3.9} & \colorbox{increase}{\footnotesize +3.9} & \colorbox{increase}{\footnotesize +1.7} & \colorbox{increase}{\footnotesize +7.6} \\
      \midrule
      \multirow{5}[0]{*}{ERNIE} & Sentence Pair & 77.5 & 75.7 & 75.0 & 56.9 & 83.3 & 28.7 & 48.9 & 66.8 \\
         & \multirow{2}[0]{*}{Flatten} & 78.6 & 76.7 & 75.8 & 56.6 & 83.8 & 30.9 & 41.0 & 70.8 \\
         &    & \colorbox{increase}{\footnotesize +1.1} & \colorbox{increase}{\footnotesize +1.0} & \colorbox{increase}{\footnotesize +0.8} & \colorbox{decrease}{\footnotesize -0.3} & \colorbox{increase}{\footnotesize +0.5} & \colorbox{increase}{\footnotesize +2.2} & \colorbox{decrease}{\footnotesize -7.8} & \colorbox{increase}{\footnotesize +4.0} \\
         & \multirow{2}[0]{*}{Hierarchical} & 79.6 & 77.5 & 76.6 & 59.0 & 84.3 & 32.7 & \textbf{49.5} & 69.6 \\
         &    & \colorbox{increase}{\footnotesize +2.1} & \colorbox{increase}{\footnotesize +1.8} & \colorbox{increase}{\footnotesize +1.7} & \colorbox{increase}{\footnotesize +2.1} & \colorbox{increase}{\footnotesize +1.1} & \colorbox{increase}{\footnotesize +4.0} & \colorbox{increase}{\footnotesize +0.6} & \colorbox{increase}{\footnotesize +2.8} \\
      \bottomrule
      \end{tabular}%
    }
    \caption{
    Experimental results.
    Performance \colorbox{increase}{increases} and \colorbox{decrease}{decreases} compared to Sentence Pair are marked.
    }
    \label{tab:results}%
    \vspace{-0mm}
\end{table*}%

\subsection{Statistics and Annotation Analysis}
\label{subsec:statistics}

Table \ref{tab:dataset} shows the statistics of \textsc{CDConv}.
It contains 11,660 conversations, where the average lengths of $u_1, b_1, u_2, b_2$ are 16.4, 12.1, 11.1, 11.6 respectively.
The ratio of positive and negative samples is 1.68 (7,309 / 4,351).
Both positive and negative samples include conversations constructed using various trigger methods, which suggests a high diversity of \textsc{CDConv}.
Among the negative samples, History Contradiction occupies the largest proportion (70.1\%) along with rich persona labels.

To shed light on the trigger methods and the chatbot behaviors, we show in Table \ref{tab:annotation} the comprehensive annotation statistics.
For the \textbf{trigger methods}, they all can effectively trigger dialogue contradictions.
Notably, Short and Inquiring (User-M) are the most effective in triggering \colorbox{role}{Role Confusion} and \colorbox{history}{History Contradiction} respectively.
For the \textbf{chatbot behaviors}, EVA and PLATO both produce fluent replies with little ethical risk, but can both be easily goaded into making contradictions.
EVA is more prone to making Intra-sentence Contradiction (\colorbox{intra_b1}{$b_1$} / \colorbox{intra_b2}{$b_2$}) and \colorbox{history}{History Contradiction}, while PLATO makes more \colorbox{role}{Role Confusion} and \colorbox{incoherent}{incoherent $b_2$}.
We speculate that their different behaviors may result from the gaps in model architectures and training corpora.

\section{Experiments}
\label{sec:experiment}

\subsection{Setups}

We randomly split \textsc{CDConv} into the training/validation/test sets with the ratio of 6/2/2.
The experiments were conducted with two settings.
The \textbf{2-class} one detects whether $b_2$ makes a contradiction, while the \textbf{4-class} one recognizes the contradiction category (the three categories in \S \ref{sec:categories} along with a non-contradiction one).
We measure model performance using \textbf{Accuracy} and \textbf{Macro-F1}.

\subsection{Compared Methods}
\label{subsec:compared}

We experimented with three popular Chinese pre-trained models: BERT, RoBERTa \cite{bert-wwm} and ERNIE \cite{ernie}.
They all contain 12 Transformer layers \cite{transformer} with the hidden size 768.
The BERT and RoBERTa are both pre-trained with whole word masking while ERNIE with the different knowledge masking strategies.
We compared three methods of contradiction detection:

\begin{figure}[t]
    \centering
    \includegraphics[width=0.85\linewidth]{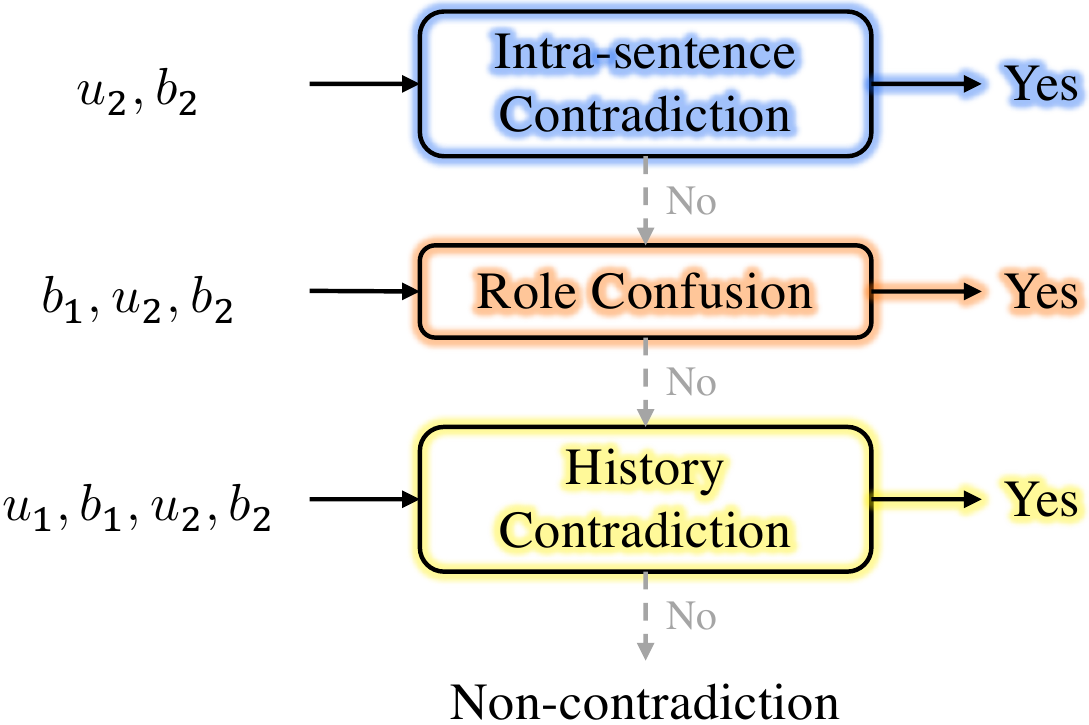}
    \caption{
    Overview of the Hierarchical method.
    }
    \label{fig:hierarchical}
    \vspace{-0mm}
\end{figure}
    
\begin{itemize}[leftmargin=*]

    \item \textbf{Sentence Pair}: The model input consists of the bot's utterances $b_1$ and $b_2$.
    This method follows the NLI framework adopted in previous work \cite{mnli, dnli, decode} where contradiction detection is performed between a pair of sentences.
    
    \item \textbf{Flatten}: The flattened whole conversation is taken as the model input, that is, $u_1, b_1, u_2$ and $b_2$.
    This method utilizes contextual information for contradiction detection in a naive way.
    
    \item \textbf{Hierarchical}: We note that the three contradiction categories are usually related to different levels of contextual information according to their definitions (\S \ref{sec:categories}).
    We thus design a hierarchical modeling method, which consists of three separately fine-tuned 2-class classifiers in sequential order (Figure \ref{fig:hierarchical}).
    Each classifier targets a specific contradiction category, takes the corresponding level of contextual information as input, and is fine-tuned with 2-class samples: the samples of the targeted contradiction category vs. all the other samples.
    Once some contradiction category is detected, it is then directly output, otherwise non-contradiction will be finally output.

\end{itemize}

In prior to fine-tuning, we pre-trained all the models on the Chinese NLI pre-training corpus, which includes two widely used Chinese NLI datasets: CMNLI \cite{clue} and OCNLI \cite{ocnli}.
We merged the ``entailment'' and ``neutral'' labels as the ``non-contradiction'' one.
See Table \ref{tab:nli} for more results of NLI pre-training.

\subsection{Implementation Details}
\label{subsec:implementation}

We implemented all experiments with the PaddlePaddle platform \cite{paddle}.
We employed the AdamW \cite{adamw} optimizer with batch size 32 and learning rate 5e-5, and used the linear learning rate scheduler with warmup proportion 0.1.
Each model was fine-tuned for 5 epochs and the checkpoint achieving the highest Macro-F1 was used for test.
We reported the average results of four random seeds, where each run took about 3 minutes on a single Tesla V100 GPU.

\begin{figure*}[t]
    \centering
    \includegraphics[width=\linewidth]{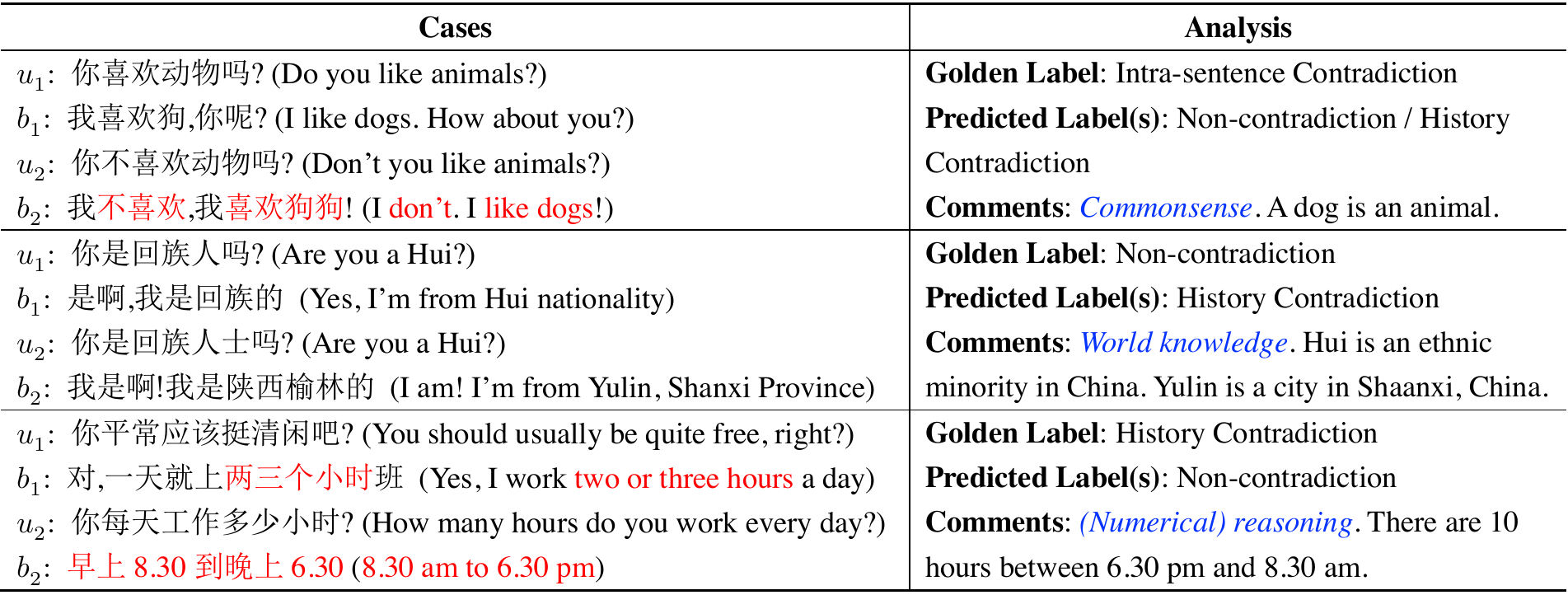}
    \caption{
    Error analysis.
    }
    \label{fig:error}
    \vspace{-0mm}
\end{figure*}

\subsection{Results}

Table \ref{tab:results} shows the results of the 2-class setting, the 4-class setting, and the fine-grained F1 scores of all the categories of the 4-class setting.
We have three major observations:

\vspace{-1mm}
\paragraph{(1) Sentence Pair performs worse than Flatten and Hierarchical.}
It is unsurprising since exploiting contextual information is critical for dialogue contradiction detection, as discussed in \S \ref{sec:intro}.

\vspace{-1mm}
\paragraph{(2) Hierarchical consistently performs best and boosts all the fine-grained results.}
Specially, Intra-sentence Contradiction and Role Confusion cannot be improved by naively feeding the models with the flattened whole conversation, see the marked \colorbox{decrease}{decreased scores}.
In contrast, Hierarchical boosts the performance in Intra-sentence Contradiction and Role Confusion and meanwhile performs well in Non-contradiction and History Contradiction.
This is because Hierarchical fully considers the characteristics of different contradiction categories and properly utilizes the required contextual information for detection.
For instance, Role Confusion needs to judge whether $b_2$ is a reply to $u_2$ or a reply to $b_1$.
It is sufficient for the classifier of Role Confusion to make use of the three utterances, while further adding $u_1$ may instead introduce noise and impair performance.

\vspace{-1mm}
\paragraph{(3) Even for Hierarchical, the performance in Intra-sentence Contradiction and Role Confusion is still poor.}
Their highest Macro-F1 are 33.0 and 49.5 respectively, which are far inferior to Non-contradiction (85.1) and History Contradiction (71.0).
One potential cause is the imbalance of samples of non-contradiction and three contradiction categories (Table \ref{tab:dataset}).
Another important reason may be that these pre-trained models still do not have a good ability of dialogue representation, which may be alleviated by additional pre-training on dialogue corpora.

\subsection{Error Analysis and Discussion}

We manually inspected the cases misclassified by the four RoBERTa Hierarchical models (trained with four random seeds).
Figure \ref{fig:error} shows the results of error analysis.
Besides proper dialogue modeling (e.g., the hierarchical way), dialogue contradiction detection also requires more abilities such as commonsense, knowledge grounding, and reasoning, which correspond to the cases in Figure \ref{fig:error}.
Though innate to human, these capabilities are still largely lacked by even gigantic deep neural models \cite{marcus2018deep, choi2022curious}.
These challenges of dialogue contradiction detection manifest that further exploration is worthy.

\section{Conclusion}

In this work, we present \textsc{CDConv}, a benchmark for contradiction detection in Chinese conversations.
By simulating common user behaviors that trigger chatbots to make contradictions, we collect 12K conversations annotated with three typical contradiction behaviors.
Experiments show that contextual information plays an important role in dialogue contradiction detection.
However, there are still unresolved challenges in \textsc{CDConv}, such as dialogue modeling, commonsense, knowledge grounding and reasoning.
We hope that \textsc{CDConv} can inspire and facilitate future research in dialogue contradiction detection and consistent generation.

\section{Ethical Considerations}

\paragraph{Human Annotation}
The human inspection and annotation was conducted by a reputable data annotation company, and the annotators are compensated fairly based on the market price.
We did not directly contact the annotators and their privacy can be well preserved.
This work does not use any demographic or identity characteristics.

\paragraph{Data Disclaimer}
In the construction of the \textsc{CDConv} conversations, the $u_1$ utterances use the dialogue posts from the open-sourced, crowd-sourced corpus DuPersona (\S \ref{subsec:procedure}).
The $u_2$ utterances either come from DuPersona or are constructed using publicly available resources (corpora, models or API, \S \ref{subsec:methods}).
The $b_1$ and $b_2$ utterances are all produced by chatbots.
Due to the potential ethical risks in these utterances, we have censored and filtered out conversations that contained unsafe or unethical contents through human inspection.

\section*{Acknowledgements}

This work was supported by the National Science Foundation for Distinguished Young Scholars (with No. 62125604) and the NSFC projects (Key project with No. 61936010 and regular project with No. 61876096). This work was also supported by the Guoqiang Institute of Tsinghua University, with Grant No. 2019GQG1 and 2020GQG0005, and sponsored by Tsinghua-Toyota Joint Research Fund.

\bibliography{custom}
\bibliographystyle{acl_natbib}

\appendix

\begin{table*}[t]
    \centering
    \scalebox{0.85}{
      \begin{tabular}{ccccc||cc}
      \toprule
      \multirow{2}[2]{*}{\textbf{Models}} & \multirow{2}[2]{*}{\textbf{Pre-training}} & \multirow{2}[2]{*}{\textbf{Fine-tuning}} & \multicolumn{2}{c}{\textbf{2-class}} & \multicolumn{2}{c}{\textbf{4-class}} \\
      \cmidrule(lr){4-5}\cmidrule(lr){6-7}
      &    &    & \textbf{Acc} & \multicolumn{1}{c}{\textbf{F1}} & \textbf{Acc} & \textbf{F1} \\
      \midrule
      \multirow{7}[2]{*}{BERT} & CMNLI & -  & 64.9 & 62.6 & -  & - \\
         & OCNLI & -  & 64.5 & 61.0 & -  & - \\
         & CMNLI + OCNLI & -  & 65.4 & 62.6 & -  & - \\
  \cmidrule{2-7}       & -  & \textsc{CDConv} & 72.3 & 70.1 & 69.2 & 51.7 \\
         & CMNLI & \textsc{CDConv} & 76.1 / \colorbox{increase}{\footnotesize +3.8} & 74.8 / \colorbox{increase}{\footnotesize +4.6} & 71.5 / \colorbox{increase}{\footnotesize +2.3} & 53.8 / \colorbox{increase}{\footnotesize +2.1} \\
         & OCNLI & \textsc{CDConv} & 74.8 / \colorbox{increase}{\footnotesize +2.5} & 72.4 / \colorbox{increase}{\footnotesize +2.3} & 72.0 / \colorbox{increase}{\footnotesize +2.7} & 52.6 / \colorbox{increase}{\footnotesize +0.9} \\
         & CMNLI + OCNLI & \textsc{CDConv} & 75.3 / \colorbox{increase}{\footnotesize +3.0} & 73.8 / \colorbox{increase}{\footnotesize +3.6} & 72.3 / \colorbox{increase}{\footnotesize +3.0} & 54.5 / \colorbox{increase}{\footnotesize +2.8} \\
      \midrule
      \multirow{7}[2]{*}{RoBERTa} & CMNLI & -  & 64.8 & 62.2 & -  & - \\
         & OCNLI & -  & 64.0 & 56.5 & -  & - \\
         & CMNLI + OCNLI & -  & 65.6 & 62.4 & -  & - \\
  \cmidrule{2-7}       & -  & \textsc{CDConv} & 72.1 & 69.9 & 69.2 & 50.7 \\
         & CMNLI & \textsc{CDConv} & 76.5 / \colorbox{increase}{\footnotesize +4.5} & 74.5 / \colorbox{increase}{\footnotesize +4.6} & 72.4 / \colorbox{increase}{\footnotesize +3.2} & 54.1 / \colorbox{increase}{\footnotesize +3.4} \\
         & OCNLI & \textsc{CDConv} & 74.1 / \colorbox{increase}{\footnotesize +2.1} & 72.4 / \colorbox{increase}{\footnotesize +2.5} & 70.6 / \colorbox{increase}{\footnotesize +1.4} & 48.5 / \colorbox{decrease}{\footnotesize -2.1} \\
         & CMNLI + OCNLI & \textsc{CDConv} & 75.7 / \colorbox{increase}{\footnotesize +3.6} & 73.7 / \colorbox{increase}{\footnotesize +3.9} & 72.2 / \colorbox{increase}{\footnotesize +3.1} & 55.1 / \colorbox{increase}{\footnotesize +4.4} \\
      \midrule
      \multirow{7}[2]{*}{ERNIE} & CMNLI & -  & 64.7 & 61.8 & -  & - \\
         & OCNLI & -  & 64.8 & 57.9 & -  & - \\
         & CMNLI + OCNLI & -  & 64.6 & 61.5 & -  & - \\
  \cmidrule{2-7}       & -  & \textsc{CDConv} & 74.3 & 72.3 & 72.4 & 54.1 \\
         & CMNLI & \textsc{CDConv} & 77.4 / \colorbox{increase}{\footnotesize +3.1} & \textbf{76.0} / \colorbox{increase}{\footnotesize +3.7} & 74.2 / \colorbox{increase}{\footnotesize +1.7} & 52.6 / \colorbox{decrease}{\footnotesize -1.5} \\
         & OCNLI & \textsc{CDConv} & 75.4 / \colorbox{increase}{\footnotesize +1.2} & 73.1 / \colorbox{increase}{\footnotesize +0.7} & 72.8 / \colorbox{increase}{\footnotesize +0.4} & 53.5 / \colorbox{decrease}{\footnotesize -0.6} \\
         & CMNLI + OCNLI & \textsc{CDConv} & \textbf{77.5} / \colorbox{increase}{\footnotesize +3.2} & 75.7 / \colorbox{increase}{\footnotesize +3.4} & \textbf{75.0} / \colorbox{increase}{\footnotesize +2.5} & \textbf{56.9} / \colorbox{increase}{\footnotesize +2.8} \\
      \bottomrule
      \end{tabular}%
      }
    \caption{
    Experimental results of NLI pre-training with the method Sentence Pair in \S \ref{subsec:compared}.
    Among the results of fine-tuning on \textsc{CDConv}, the performance \colorbox{increase}{increases} and \colorbox{decrease}{decreases} compared to no NLI pre-training are marked.
    Note that the last line of each model corresponds to the results of Sentence Pair in Table \ref{tab:results}.
    \textbf{Observation 1}: Directly applying the NLI classifiers to \textsc{CDConv} is remarkably inferior to fine-tuning.
    \textbf{Observation 2}: NLI pre-training generally leads to improvements, and using both CMNLI and OCNLI for pre-training gives the best performance under the 4-class setting.
    }
    \label{tab:nli}%
    \vspace{-0mm}
\end{table*}%

\section{Limitations}
\label{sec:limitations}

\paragraph{Data Coverage and Construction} 
An ideal benchmark for dialogue contradiction detection may be expected to (1) cover as many and diverse contradiction cases as possible, and (2) be close to the real scenario of human-bot conversation scenario.
However, the cases of non-contradiction and contradiction in natural human-bot conversations are extremely unbalanced, as stated in \S \ref{sec:categories} and \cite{decode}, which brings great difficulty for the data collection.
For this reason, we (1) focus on the three typical contradiction categories in the manually inspected contradiction cases (\S \ref{sec:categories}), and (2) construct conversations by simulating common user behaviors that trigger contradictions.

We are explicitly aware that \textsc{CDConv} has a finite coverage of the cases of dialogue contradiction.
Specially, the \textsc{CDConv} conversations consist of only two turns, but (1) contradictions may occur after more than one turns, and (2) some contradiction cases, especially History Contradiction, may contradict multiple turns.
The samples of (1) can be obtained by applying data augmentation to the \textsc{CDConv} conversations based on chatbots' self-chat \cite{eva2.0, platoxl} or language models' completion \cite{zheng2022augesc, dai2022dialog}.
The samples of (2) are not covered by \textsc{CDConv} but in fact rarely occur based on our observations.
Future benchmarks for dialogue contradiction detection may consider these complex cases of (2).

\vspace{-1mm}
\paragraph{Fluency and Coherence of Conversations} 
From Table \ref{tab:annotation}, we observed that Inquiring (User) results in more \colorbox{incoherent}{incoherent $b_2$}.
The three Perturbation methods also lead to more \colorbox{non-fluent}{non-fluent $u_2$}.
It indicates that these methods may somewhat impair the naturalness of conversations.
To address this, we conducted elaborated manual inspection (the 3rd and 4nd steps in \S \ref{subsec:procedure}) to filter out the conversations containing non-fluent or incoherent replies.

\vspace{-1mm}
\paragraph{Human Annotation} 
Due to the subjectivity of human annotation, there may unavoidably exist mislabeled samples in \textsc{CDConv}.
To alleviate this, we have adopted the mode of multi-person annotation, conducted spot check for each annotation batch, and required the pass rates to reach 95\% to ensure data quality (\S \ref{subsec:quality}).
We especially point out that, despite the mode of multi-person annotation, there may still exist biases in the annotation results regarding ``fluency'' (\S \ref{subsec:procedure}).
Due to the characteristics of the Chinese language and the difference in individual habits of language usage and communication, the annotators' understanding of ``fluency'' may not be identical.
Although we have tried our best to unify the annotation criteria through constant feedback and quality check (\S \ref{subsec:quality}), these biases may not be eliminated completely.

\section{Definitions of Persona Labels}
\label{sec:persona_definition}

\begin{itemize}[leftmargin=*]
\setlength\itemsep{-1mm}

    \item \textbf{Persona Attributes}:
    The properties of the speakers and their relationships, including but not limited to: name, gender, age and date of birth, occupation and salary, residence place, family members, belongings (e.g., pets, cars, houses), etc.
    
    \item \textbf{Persona Opinions}:
    The speakers' preferences and opinions on other people or things, including but not limited to: hobbies, preferences, opinions on animals, food, movies, books, music, etc.
    
    \item \textbf{Persona Experiences}:
    Past, present or future events experienced by the speakers.
    
    \item \textbf{Persona-unrelated}:
    Other information involved in History Contradiction (e.g., named entities, world knowledge or facts).

\end{itemize}

\end{document}